\documentclass{article}
\usepackage{makecell}
\usepackage{subfig}
\usepackage[algo2e]{algorithm2e}
\usepackage[export]{adjustbox}
\usepackage{hyperref}
\usepackage{multirow}
\hypersetup{hidelinks,
	colorlinks=true,
	allcolors=black,
	pdfstartview=Fit,
	breaklinks=true}
\usepackage{ijcai24}

\usepackage{times}
\usepackage{soul}
\usepackage{url}
\usepackage{hyperref}
\usepackage[utf8]{inputenc}
\usepackage{caption}
\usepackage{graphicx}
\usepackage{amsmath}
\usepackage{amsthm}
\usepackage{booktabs}
\usepackage{algorithm}
\usepackage{algorithmic}
\usepackage[switch]{lineno}
\usepackage{listings}
\usepackage{xcolor}

\title{MEIA: Multimodal Embodied Perception and Interaction in Unknown Environments}

\author{
Yang Liu$^1$\and
Xinshuai Song$^1$\and
Kaixuan Jiang$^1$\and
Weixing Chen$^1$\and
Jingzhou Luo$^1$\and
Guanbin Li$^1$\And
Liang Lin$^1$\\
\affiliations
$^1$Sun Yat-sen University\\
\emails
\ liuy856@mail.sysu.edu.cn,
songxsh@mail2.sysu.edu.cn,
 jiangkx3@mail2.sysu.edu.cn,
 chen867820261@gmail.com,
 luojzh5@mail2.sysu.edu.cn,
 liguanbin@mail.sysu.edu.cn,
 linliang@ieee.org
}

\begin{document}

\maketitle

\begin{figure*}[!t]
	\centering
    \includegraphics[width=1\textwidth]{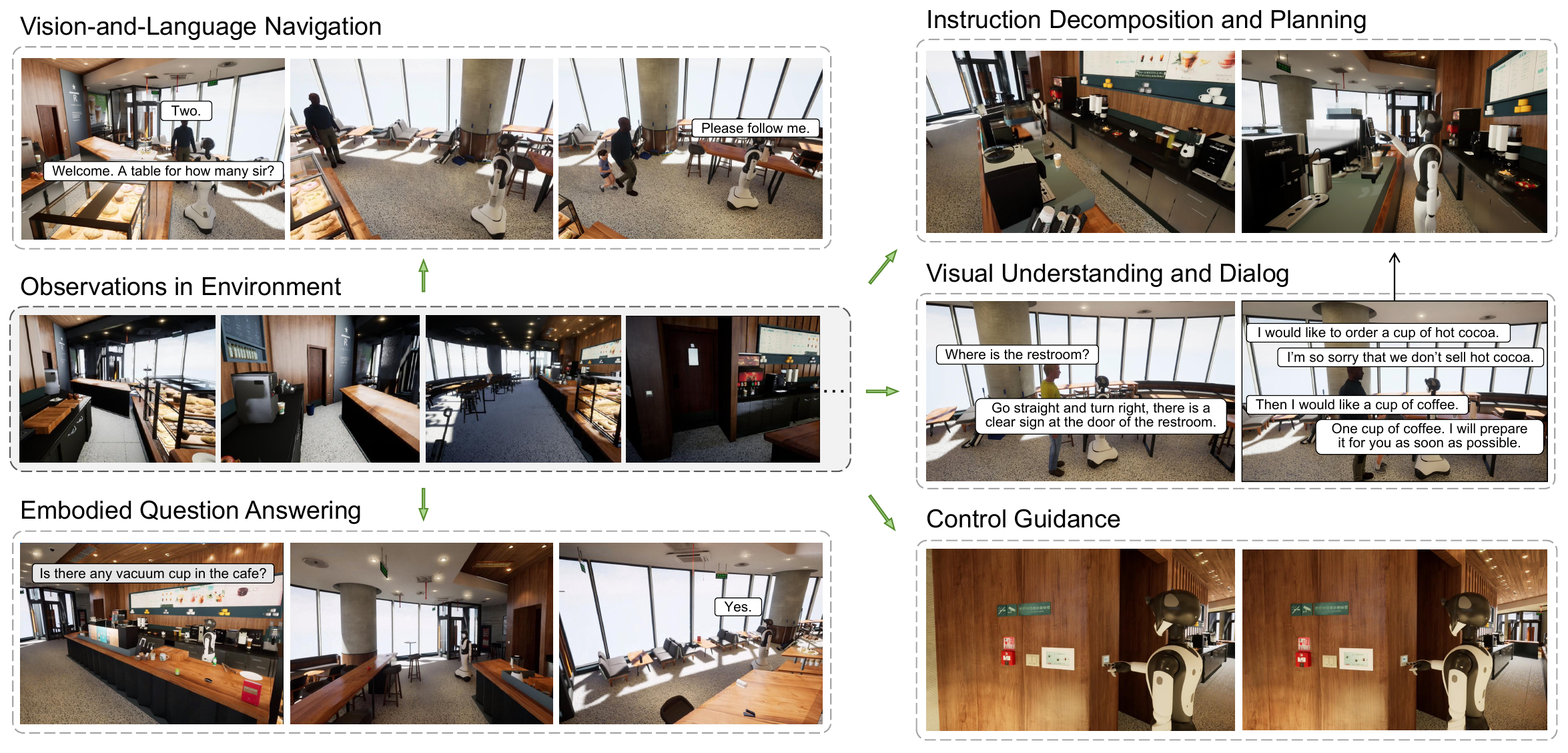}
	\caption{We propose MEIA, a model that decomposes high-level language instructions into a series of executable actions.}
	\label{fig:task}
\end{figure*}

\begin{abstract}
With the surge in the development of large language models, embodied intelligence has attracted increasing attention. Nevertheless, prior works on embodied intelligence typically encode scene or historical memory in an unimodal manner, either visual or linguistic, which complicates the alignment of the model's action planning with embodied control. To overcome this limitation, we introduce the Multimodal Embodied Interactive Agent (MEIA), capable of translating high-level tasks expressed in natural language into a sequence of executable actions. Specifically, we propose a novel Multimodal Environment Memory (MEM) module, facilitating the integration of embodied control with large models through the visual-language memory of scenes. This capability enables MEIA to generate executable action plans based on diverse requirements and the robot’s capabilities.
Furthermore, we construct an embodied question answering dataset based on a dynamic virtual cafe environment with the help of the large language model. In this virtual environment, we conduct several experiments, utilizing multiple large models through zero-shot learning, and carefully design scenarios for various situations.
The experimental results showcase the promising performance of our MEIA in various embodied interactive tasks.
\end{abstract}


\section{Introduction}
\label{sec:intro}

Picture yourself entering a coffee shop staffed by robots. Upon recognizing your arrival and that of your friends, a robot approaches, greets you, and guides you to suitable seats based on the group size. Subsequently, it proceeds to the bar, takes your coffee and dessert orders, and then prepares your order. Meanwhile, you request the robot to adjust the environment by lowering the curtains and decreasing the air conditioning due to the rising outside temperature and warmth inside the cafe. Additionally, it is able to respond to questions you asked based on the environment.
This scenario illustrates the agent's responsibilities in vision-and-language navigation, instruction decomposition and planning, as depicted in Fig.\ref{fig:task}. These tasks demand diverse capabilities,  including visual recognition \cite{liu2018transferable,liu2023self,yan2023skeletonmae}, target navigation \cite{liu2022causal}, language comprehension \cite{liu2023cross,10.1145/3581783.3611873}, commonsense reasoning \cite{liu2023causalvlr,chen2023visual}, task planning \cite{zhu2022hybrid,wang2023urban,ijcai2023p672}, and object manipulation \cite{tang2023towards}. Termed embodied AI, it stands as a prominent research focus in the realms of artificial intelligence and robotics. Unlike conventional AI learning from datasets comprised mainly of internet-curated images \cite{liu2016combining,liu2019deep,xu2022dual}, videos \cite{liu2022tcgl,liu2021semantics,liu2018hierarchically,liu2018global}, or text \cite{liu2023vcd}, embodied AI involves learning from self-centered data, akin to human experiences \cite{duan2022survey}. Agents must collect vital visual information and engage with the physical world based on real-time observations in embodied AI tasks \cite{liu2024aligning}.

In recent years, large language models (LLMs) and vision language models (VLMs) have received widespread attention. Leveraging massive training data on the web, LLMs such as GPT-4 \cite{openai2023gpt4}, PaLM \cite{chowdhery2023palm}, and LLaMA \cite{touvron2023llama} have demonstrated powerful capabilities in tasks like dialogue and reasoning. However, the challenge arises when LLMs attempt simple planning and reasoning in the physical world as the lack of physical grounding. Some works \cite{driess2023palm,singh2023progprompt,song2023llm} attempt to integrate visual information into LLMs, utilizing LLMs to execute embodied tasks and resulting in good performance. Other works \cite{hu2023look} choose to employ VLMs for embodied tasks, which can accomplish tasks by directly integrating observation images of the environment into reasoning and planning processes, showing impressive performance in zero-shot scene understanding and reasoning. However, several issues remain: 1) how to better store environmental information to provide richer content for large models, and 2) whether LLMs and VLMs can work collaboratively to solve embodied tasks.

To address the above challenges, we propose a novel embodied model MEIA based on LLMs and VLMs, which excels in performing a variety of embodied AI tasks such as embodied question answering and embodied control in a physical environment. Our MEIA comprises three essential modules that work collaboratively, including i) vision module utilized to generate environmental language memory and environmental image memory based on the surroundings, ii) control module used to dialogue with guests and perform embodied operations such as seating guidance and cleaning up, iii) large model responsible for reasoning and action planning. Since a good environmental memory storage method aids agents in remembering the environment and making correct decisions to facilitate task completion, we propose a novel Multi-modal Environmental Memory (MEM) module incorporating object descriptions and floor plan of the environment. To further enhance the accuracy and feasibility of model output, we meticulously designed prompts to guide the LLMs and VLMs.
In addition, we also build an embodied question answering dataset with a thousand pieces of data based on the simulator provided by Dataa Robotics, which provides a cafe scene that is close to reality and achieves abundant human-machine interaction design.
We evaluate our MEIA in the simulator and on the dataset we constructed.
The contributions can be summarized as follows:
\begin{itemize}
  \item We introduce an embodied model MEIA that seamlessly integrates large language models and vision language models. Our MEIA can translate high-level tasks into executable action sequences.
  \item We propose a Multi-modal Environment Memory (MEM) module that equips the MEIA with more detailed embodied knowledge, enhancing the understanding of the physical environment and experience.
  \item We utilize a large language model to effectively generate data with high quality, constructing an embodied question answering dataset.
  \item Our MEIA exhibits robust zero-shot learning capabilities and is capable of completing the embodied task with good performance.
\end{itemize}

\section{Related Work}

\subsection{Vision Language Model}

The VLMs excel in simultaneous comprehension of both images and text, which can be used in tasks such as caption generation, visual question answering and commonsense reasoning. Like UNITER \cite{chen2020uniter} and BLIP \cite{li2022blip}, previous works mostly rely on neural networks such as visual transformer (ViT) and recurrent neural network (RNN), to construct vision language models. These models utilize self-attention mechanisms to establish alignment between text and images. Inspired by the success of large language models, there has been a growing interest in constructing large-scale vision language models. Some works \cite{li2023blip,dai2023instructblip,liu2023visual} combine pre-trained LLMs with visual reasoning, leveraging visual models. In this work, we select GPT-4-Vision as our vision language model due to its superior performance.

\subsection{Memory of Environment}
The retention of environmental observations is crucial in embodied tasks, as the absence of grounding in the physical environment might lead to inaccurate responses and unfeasible planning for LLMs and VLMs. LLM-Planner \cite{song2023llm}, VELMA \cite{schumann2023velma}, NavGPT \cite{zhou2023navgpt} and LLM-Grounder \cite{yang2023llm} take the textual descriptions of visual observations as input or attach them to prompts. However, pure text environmental memory will lose a great deal of detail because of the constraints of text description. Some works focus on storing environmental information through graphs constructed in various forms, such as 2D top-down maps \cite{min2021film,inoue2022prompter}, 3D voxel maps \cite{tan2023knowledge,murray2022following}, 3D maps \cite{hong20233d}, etc. Inspired by these works, we introduce a novel memory module: the Multi-modal Environmental Memory (MEM), including object IDs and coordinates as environmental language memory and visual observation pictures as environmental image memory.

\subsection{Embodied AI Tasks}
In embodied AI, agents learn through interactions with environments from an egocentric perception similar to humans \cite{duan2022survey}. Embodied AI tasks mainly encompass visual exploration and navigation, embodied planning, embodied control, and embodied question answering. As the most intricate task, embodied question answering \cite{das2018embodied} requires the agent to answer questions related to the physical world based on information obtained from visual exploration and navigation. Additionally, the agent needs to interact with the environment through embodied planning and embodied control to obtain more information. Unlike existing works \cite{driess2023palm,singh2023progprompt,hu2023look,ding2023leveraging,yaoxian2023scene} that utilize either LLMs or VLMs to solve these tasks, we seamlessly integrate both LLMs and VLMs. We leverage LLMs to understand customers’ needs, and decompose and plan tasks expressed in language. Simultaneously, VLMs aid in achieving environmental space understanding, location navigation, robot motion control, and embodied question answering.


\section{Simulation Configuration}

Our MEIA is performed in the simulator provided by \href{https://www.dataarobotics.com/en}{\emph{Dataa Robotics}}, which is built on the UE5 and provides a cafe human-computer interaction scene with rich details. In this section, we introduce the configuration of the scene and robot.

\subsection{Scene Configuration}

In terms of scene configuration, the simulator offers a cafe environment designed to closely emulate reality. Distinguishing itself from other simulators like Habitat \cite{habitat}, it uses high-resolution models as well as precise materials and textures, ensuring an accurate representation of object dynamics and mechanical behavior through a sophisticated physics engine and making the appearance and physical characteristics of objects in the simulator environment very close to the real world. In addition, the simulator provides a robot control system and object interaction to achieve a rich human-machine interaction design.

The various objects in the simulator provide strong support for robot navigation and operation training. In the cafe scene, there are a total of \textbf{13 categories} of nearly \textbf{800 objects}, including interactive objects such as coffee machines, kettles, floor scrubbers, air conditioners, light switches, etc. Additionally, \textbf{73 categories} of objects can be added. The cafe scene can be seen in Fig.\ref{fig:task}.


\subsection{Robot Configuration}

The robot in the simulation environment are designed based on \href{https://www.dataarobotics.com/en/product-44.html}{\emph{Dataa Robotics’ Ginger humanoid robot intelligent agent}}. It has a feature-rich environment interaction interface and supports robot joint control, chassis speed control, and multiple sensor simulations.

The visual capabilities of the robot include:



\begin{itemize}
  \item Head, chest, and waist cameras capture \textbf{RGB images},
  \item Head, chest, and waist cameras capture \textbf{depth images},
  \item Head camera captures \textbf{segmentation images}.
\end{itemize}

The control capabilities of the robot include:





\begin{itemize}
  \item Precise control of 21 joints in the robot’s torso and arms, and 10 joints in its hands.
  \item Precise movement to specified coordinates.
  \item Joint movement control (IK control) to specified coordinates.
  \item The ability to grasp, suction, and release cylindrical objects.
\end{itemize}

In addition, the robot can display the content of the conversation.
It can also perform complex interactive operations such as turning on/off lights, moving tables and chairs, pushing carts, using vacuum cleaners, using coffee machines, and operating air conditioning control panels, as shown in Fig.\ref{fig:task}.

\begin{figure*}[!t]
  \centering
  \includegraphics[width=1\textwidth]{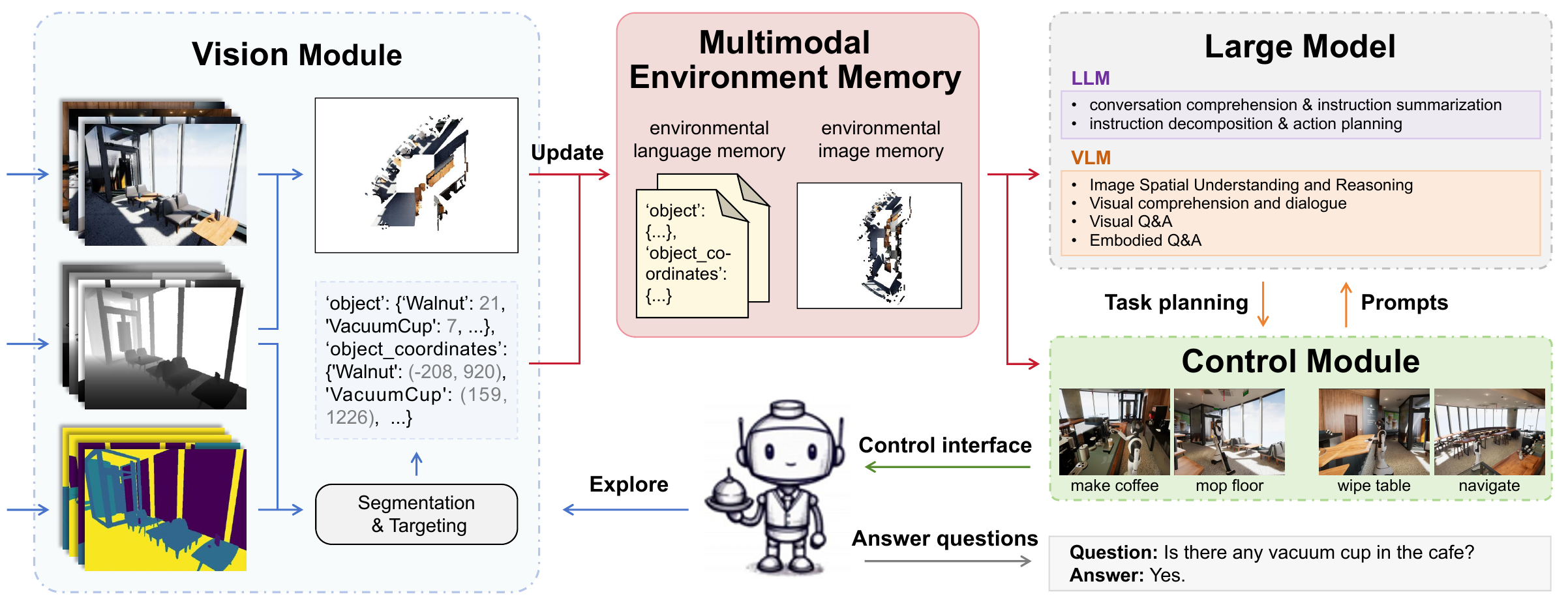}
    \vspace{-10pt}
  \caption{The structure diagram of MEIA. MEIA is implemented through three functional modules: the vision module, the control module, and the large model. The multimodal environmental memory generated by the vision module will serve as a bridge between the control module and the large model, enabling them to work collaboratively to complete tasks, and achieving efficient integration of large model perception, memory, and embodied control.}
    \vspace{-10pt}
  \label{fig:model}
\end{figure*}

\section{MEIA: An Agent for Multimodal Embodied Interaction and Control}

The structure of MEIA is shown in Fig.\ref{fig:model}. MEIA consists of three modules: the vision module, the control module, and the large model. The multimodal memory obtained by the vision module is the core of MEIA. It serves as a bridge between the control module and the large model, enabling the large model to generate highly executable action sequences for diverse needs and hand them over to the control module for execution. The details of MEIA is as follows.


\subsection{Multimodal Environment Memory}

The embodied control and the large model cannot be directly interacted due to the domain gap. The integration of embodied control and the large model has always been the focus of embodied AI. Previous works \cite{song2023llm,zhou2023navgpt} usually implement memory of scenes or historical trajectories in a single-modal form of visual or natural language modality, making the plans given by the large model difficult to execute by embodied agents. Recently, some works attempt to delineate the scope of action plans generated by large models to achieve the executability of the plans \cite{prompt_gpt}. To address this issue, we propose the multimodal environment memory (MEM) module (Fig.\ref{fig:model}), which stores the names and coordinates of key items in the scene recorded in natural language and the two-dimensional plan of the environment as memory,
serving as a guidance for the large model to achieve highly executable action plans under diverse needs. The details of the construction of MEM are as follows.

\textbf{Environmental Language Memory.} The visual interface of the robot can obtain different modalities of images, including RGB images, depth images, and segmentation images.
Segmentation images set the pixels corresponding to different key items in the scene to their ID values to achieve item segmentation. We use the K-Means algorithm to calculate the pixel coordinates of the key items. The algorithm is shown in Algorithm \ref{alg:center}.

After obtaining the pixel coordinates, we calculate the camera coordinates. With camera intrinsic parameters denoted as $f_x, f_y, c_x, c_y$, and given object pixel coordinates $(i, j)$ along with the corresponding object depth at that pixel, represented by $depth$, the calculation proceeds as follows:
\begin{align}
    \begin{cases}
    \begin{aligned}
    x_c &= \frac{(i-c_x)}{f_x}*depth\\
    y_c &= \frac{(j-c_y)}{f_y}*depth\\
    z_c &= depth
    \end{aligned}
    \end{cases}
\end{align}

\begin{algorithm}[!t]
    \caption{Center positioning algorithm}
    \label{alg:center}
    \textbf{Input}: Pixel coordinate set list $C$, threshold $\zeta$\\
    \textbf{Output}: Pixel coordinate center $c$
    \begin{algorithmic}[1] 
        \STATE Initialize \textbf{K-Means} algorithm $cls$, set number of cluster centers $n \_clusters=2$.
        \STATE Calculate cluster centers $cluster = clf.fit(C)$
        \STATE Calculate the distance between cluster centers \\
        $d = \sqrt{\sum_{i=1}^2(cluster_0[i]-cluster_1[i])^2} $.
        \IF {$d>\zeta$}
        \STATE $c = cluster[0]$.
        \ELSE
        \STATE $c = (cluster[0] + cluster[1])/2$.
        \ENDIF
    \end{algorithmic}
\end{algorithm}
\vspace{-5pt}

Given the camera coordinates $(x_c, y_c, z_c)$, for the camera extrinsic parameters, we have a rotation matrix $R$ and a translation vector $T$:
\begin{align}
    (x_g, y_g, z_g) = R*[x_c, y_c, z_c] + T
\end{align}

When obtaining the coordinates $(x_g, y_g, z_g)$ of the ginger coordinate system, it should be noted that the ginger coordinate system is opposite to the x-axis of the world coordinate system. Additionally, it has Euler angles $\theta_g$ and a translation vector $T_g$.

The rotation matrix $R_g$ is obtained by calculating the Euler angles:

\begin{align}
    R_x = \begin{bmatrix}
            1 & 0 & 0\\
        0 & cos \alpha & -sin \alpha\\
        0 & sin \alpha & cos \alpha
\end{bmatrix}\\
R_y = \begin{bmatrix}
            cos \beta & 0 & sin \beta\\
        0 & 1 & 0\\
        -sin \beta & 0 & cos \beta
\end{bmatrix}\\
R_z = \begin{bmatrix}
            cos \gamma & -sin \gamma & 0\\
        sin \gamma & cos \gamma & 0\\
        0 & 0 & 1
\end{bmatrix}
\end{align}
\begin{align}
    R_g = R_z*R_y*R_x
\end{align}

Finally, the world coordinates are obtained:
\begin{align}
    (x_w, y_w, z_w) = R_g*[x_g, -y_g, z_g] + T_g
\end{align}

\textbf{Environmental Image Memory.} The environmental image memory is an environmental plane map obtained by further projecting the 3D point cloud image constructed by the robot using the visual information gathered during the exploration process.

The robot obtains environmental visual information by calling the RGBD camera to capture RGB images and depth images. For the RGB image and depth image corresponding to the same viewing angle, with the help of Open3D, an open-source library that supports rapid processing of 3D data, a three-dimensional point cloud with color information is constructed based on the internal parameter information of the camera. However, directly superimposing 3D point clouds obtained from multiple sets of RGBD images will cause all point cloud information to overlap and block each other. Therefore, fusing the 3D point clouds obtained from each set of images also requires calculations similar to the world coordinates of the objects, applying rotation and translation operations based on the camera's posture.

Additionally, outliers in the point cloud are identified and removed based on statistical analysis. For each point in the point cloud, we select the closest $n$ points and calculate their average distance from the point. Simultaneously, we calculate the global average distance $dis_g$ and global standard deviation $std_g$, which represent the average distance from each point in the point cloud to any other point and the standard deviation of these distances, respectively. If a point's average distance is greater than $dis_g + std_r*std_g$, it is considered an outlier, where $std_r$ specifies the multiple of the standard deviation.

\begin{figure}[!t]
    \centering
    \includegraphics[width=0.5\textwidth]{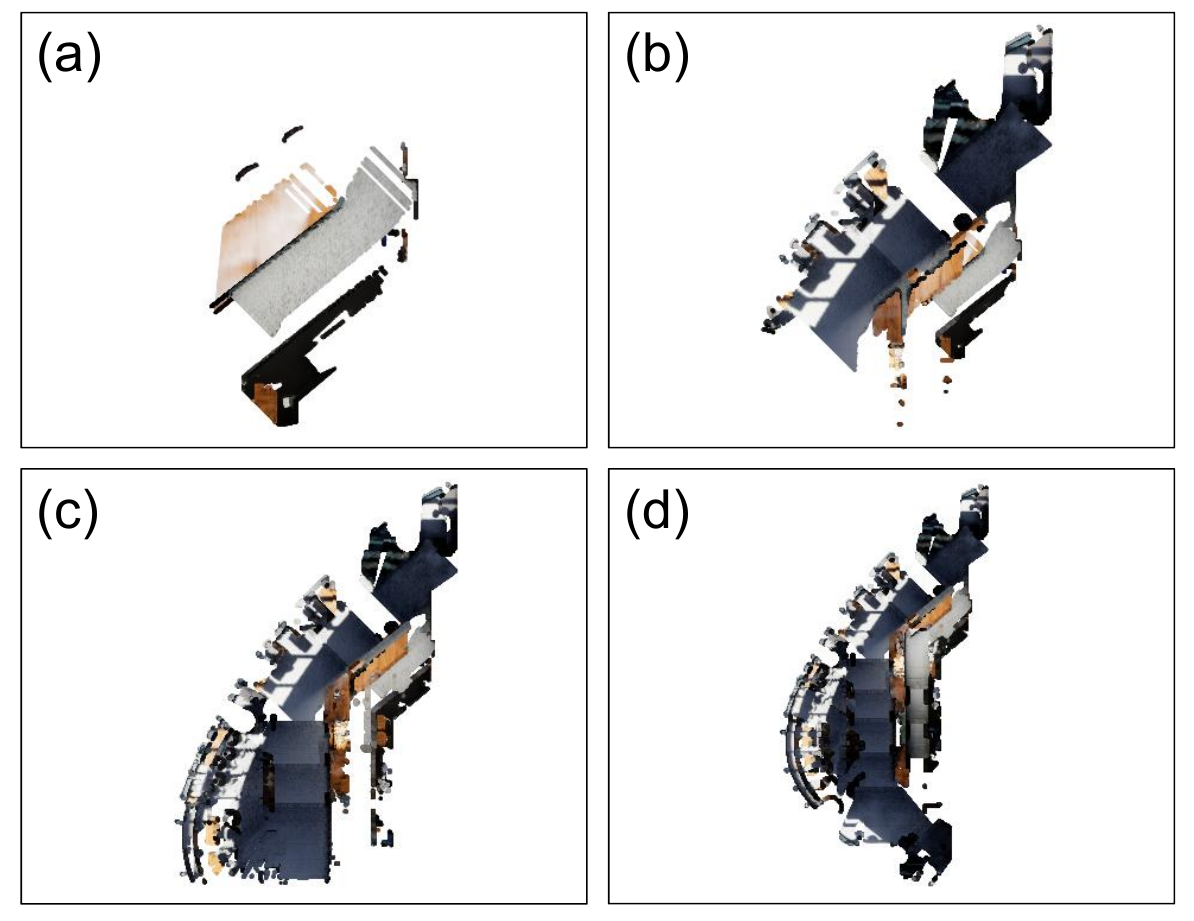}
    \vspace{-10pt}
    \caption{Construction process of an environmental floor plan.
}
    \vspace{-10pt}
    \label{fig:plan}
\end{figure}

Although the 3D point cloud contains richer and more accurate information than 2D images, processing and analyzing three-dimensional images directly can be challenging for large models. As a result, we project the 3D point cloud onto the plane where the floor is located. The environmental plane map can to some extent display the exploration status of the current environment, providing an overall understanding of the cafe's environment and determining whether the environment has been fully explored. As the exploration progresses, the 3D point cloud is enriched and the environmental plane map contains more information, as shown in Fig.\ref{fig:plan}.

By combining object IDs and coordinates with environmental plane map, we obtain multimodal environment memory. The MEM is designed as the global memory so that the robot can update multimodal memory when it calls the visual interface, to achieve flexible response for complex environment.


\subsection{Embodied Control}
Embodied control is the fundamental ability of robots to complete tasks, and its design is determined jointly by the robot control module and task requirements. We encapsulate the robot’s capabilities and design flexible control interfaces:
\begin{itemize}
  \item Move to target: specify the name of the item and move to the item based on environmental memory.
  \item Produce and grab milk: generate a glass of milk on the bar table, recognize and guide by the vision module to grab the milk.
  \item Make coffee: operate the coffee machine to make coffee.
  \item Pour water: operate the kettle to pour water.
  \item Grab bread: grab bread from the cabinet.
  \item Control air conditioning: control the air conditioning according to visual feedback, guided by VLM.
  \item Mop the floor: use a mop to mop the floor.
  \item Wipe the table: wipe the table with a towel.
  \item Control curtains: control the curtains to open or close.
  \item Control lighting: control the lighting to turn on or off.
  \item Straighten the chair: straighten the misplaced chair.
\end{itemize}












Our design for embodied control capability is versatile, allowing the robot to possess a range of abilities. It is not restricted to a single type of skill but is equipped to successfully accomplish a variety of tasks in complex scenarios.
Our embodied control module not only provides ``tools” for the large model but also integrates multimodal environmental memory to execute the action sequence planned by the large model to complete the task.

\subsection{Embodied Question Answering}
\label{section:Embodied Question Answering}
\textbf{Dataset.} We construct a new dataset based on the simulator. First, different scenarios are constructed using the GPT-4 Turbo, with several reference examples provided. Under the guidance of prompt, the GPT-4 Turbo randomly combines the objext IDs and the location coordinates within the allowed range to produce an output. According to the output, the generation function is further called to generate corresponding items in the environment, achieving the effect of random environment construction. Next, for different types of questions, templates are designed to facilitate the large-scale generation of questions. The templates are shown in Table \ref{table:template}.
Then, based on the randomly constructed environment, prompt is used to instruct the GPT-4 Turbo to generate questions and corresponding answers, which are combined with the scenes and question types as a piece of data in the dataset. For each scenario, three question-answer pairs are generated for each template. The dataset contains 70 generated scenes, with a total of 1050 pieces of data, all recorded in a json file.
Ultimately, we conduct manual inspection and correction of the data, and ensure data balance by controlling the distribution of answers in the dataset.

\begin{table}[h]
\centering
\scalebox{0.73}{
\begin{tabular}{c|c}
\hline
type                       & template                                                            \\ \hline
\multirow{2}{*}{location}  & \emph{What is the item on the same table as the $<obj>$?}                \\
                           & \emph{Are the $<obj1>$ and the $<obj2>$ on the same table?}                \\ \hline
comparing                  & \emph{Is the $<obj1>$ closer to the $<obj2>$ than to the $<obj3>$?}          \\ \hline
\multirow{2}{*}{existence} & \emph{Is there any $<obj>$ in the cafe?}                                 \\
                           & \emph{Is there anything in the cafe that I can use to $<do$ $something>$?} \\ \hline
\end{tabular}}
\caption{Question Templates}
\label{table:template}
\end{table}

\textbf{Embodied Question Answering.}
Questions, multimodal environment memories, and newly explored information serve as inputs to the multimodal large language model GPT-4 Turbo with vision. Upon reaching a location, the robot observes the environment from four directions (front, rear, left, and right) to gather new information. Acknowledging that the plans generated by the model may not always be feasible, previously unsuccessful plans are also fed back into the model to prevent the repetition of similar invalid plans. Prompted by the designed queries, GPT-4 Turbo with vision first assesses whether the existing information is adequate for providing an accurate answer to the question. In cases of insufficiency, the model devises action plans to direct the robot in moving and interacting to acquire new observations. Then, it generates output based on the new input, thereby creating a closed loop. If the existing information is sufficient, the model synthesizes the available information to answer the question, concluding the closed loop.

\subsection{Large Model}
Large model is the command center of MEIA. As shown in Fig.\ref{fig:model}, we comprehensively utilize LLMs and VLMs, combining different large models for various tasks to fully leverage the strengths of each model.

\textbf{LLM.}  For LLM, we use \textbf{GPT-3.5 Turbo} and \textbf{GPT-4 Turbo}. GPT-3.5 Turbo is used to summarize and generate natural language instructions for the conversation between the robot and the guest, as this task is relatively easy.
On the other hand, GPT-4 Turbo is employed for the intricate task of decomposing and planning instructions,
demanding robust comprehension, planning, and reasoning abilities from the LLM.

\textbf{VLM.}  GPT-4 Turbo with vision has superior capability in both visual and language understanding compared to open-source models. In our experiments, it can form spatial understanding of complex images and offer independent judgments, thereby generating content of high quality. Therefore, we use GPT-4 Turbo with vision to implement guest seating guidance task (visual language navigation), spatial-aware dialogue interaction task (understanding the space inside the cafe and completing dialogue requirements), and control guidance task (determining which action to execute). Furthermore, we also use the model for autonomous decision-making to solve embodied question answering problems, as mentioned in Section \ref{section:Embodied Question Answering}.

Inspired by the spirit of Chain-of-Thought (CoT) \cite{react} and In-Context Learning (ICL) \cite{icl}, we design specific prompts to make different large models provide accurate and appropriate solutions for different situations. The detailed prompt design can be seen in the appendix. By assigning different sub-tasks to different large models, the large models can decompose and plan various tasks based on the control module’s capabilities and multimodal environmental memory, enabling the robot to interact with guests like humans and meet their requirements.

\begin{figure*}[!t]
	\centering
    \includegraphics[width=1\textwidth]{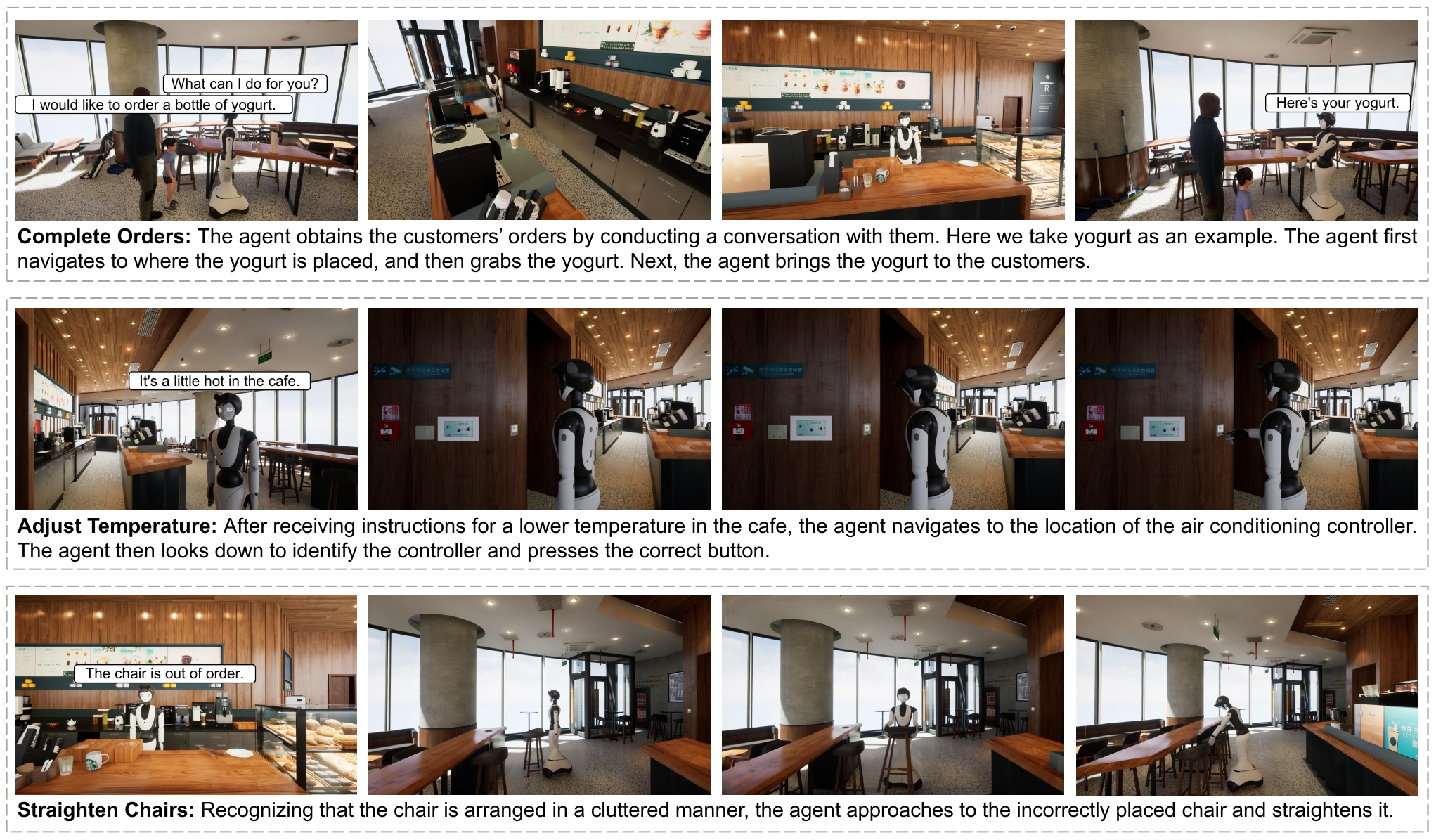}
    \vspace{-10pt}
	\caption{\textbf{Task execution.}  We present three tasks, each of which has a brief description of the implementation process.
}
\vspace{-10pt}
	\label{fig:execution}
\end{figure*}

\section{Experimental Results}
We design a plot pipeline for our solution based on reality, systematically evaluating and comparing various models for different tasks. Additionally, we also test the performance of the model on the embodied question answering dataset we construct. The details are outlined below.

\subsection{Experimental Settings}
According to the real interaction between the waiter and the customer in a cafe, we design a closed-loop plot pipeline.

In the closed-loop plot pipeline, the robot’s tasks are divided into three parts:

\textbf{Environmental Exploration}: The robot explores the environment and forms an initial multimodal environmental memory.

\textbf{Human-Robot Interaction}: The robot chooses an appropriate location to wait for guests, interacts with the guests, answers their questions, and understands their needs.

\textbf{Task Understanding and Execution}: The robot comprehends the guests' requirements, forms an action plan accordingly, and executes tasks to fulfill the guests' needs. During this period, the robot agent may encounter abnormal prompts (such as garbage on the ground), and it needs to solve the abnormal situation. We illustrate three tasks in Fig.\ref{fig:execution}.

Once the customer’s needs are all fulfilled, the agent transitions back to the Human-Robot Interaction task again. In addition, we use GPT-3.5 to randomly sample from executable subtasks and generate colloquial instructions to simulate authentic human-machine interaction scenarios, serving as evaluation cases to assess the performance of different models.

\begin{table}[!t]
\centering
\label{tab:cap}\small
\begin{tabular}{c|c|c}
  \hline
  \thead{\textbf{Model}} & \thead{\textbf{Visual}\\ \textbf{question}\\ \textbf{answering}} & \thead{\textbf{Spatial} \\ \textbf{understanding}\\ \textbf{and reasoning}}
  \\
  \hline
  miniGPT4 \cite{zhu2023minigpt} & 44\% & 8\% \\
  CogVLM \cite{wang2023cogvlm} & 52\% & 38\% \\
  CogAgent \cite{CogAgent} & 46\% & 32\% \\
  GPT4-V \cite{openai2023gpt4V(ision)} & \textbf{98\%} & \textbf{94\%} \\
  \hline
\end{tabular}
\caption{Sub-task Evaluation.}
\label{table:sub_t}
\centering
\end{table}

\subsection{Results and Comparison}
We add state-of-the-art models and methods \cite{zhu2023minigpt,wang2023cogvlm,LMPlanner,RobotPrompt} to our experimental comparisons. The outputs of large models are obtained through zero-shot learning via prompt engineering. Specific experimental details can be found in the appendix we submitted.

\textbf{Metrics.}  We use \textbf{Executable Success Rate (ESR)} to judge the performance of the models, which formulated as:
\begin{align}
    ESR = \frac{n_e}{N}
\end{align}
Where the $n_e$ is the number of successful executions and $N$ is the total number of tasks. When the output or plan given by the model can be successfully executed and complete the task in the simulator, it is counted as a success. Besides, we use \textbf{Success Rate Weighted by Step Length (SSL)} to measure the accuracy and efficiency of the plans given by different models and methods:
\begin{align}
    SSL = \frac{1}{N}\sum^N{s_i*\frac{l_c}{\max(l_g, l_p)}}
\end{align}
where $N$ is the total number of tasks, $s_i$ is the success rate of planning execution, $l_g$ is the length of the grounding planning, $l_p$ is the length of the planning generated by large models, and $l_c$ is the length of the correct step in the planning.
\subsubsection{Sub-task Evaluation}

The design of sub-task evaluation is based on the abilities required by robot agent in experimental plots, including visual question answering, and spatial understanding and reasoning. We use the latest large visual language models, miniGPT4 \cite{zhu2023minigpt}, CogVLM \cite{wang2023cogvlm} and CogAgent \cite{CogAgent} for the sub-task evaluation. For each task, we evaluate each model 50 times.

\begin{table*}[!t]\small
    \centering
    \tiny
    \adjustbox{max width=0.85\linewidth,width=0.8\linewidth,center=\linewidth}{
    \begin{tabular}{c|ccc|ccc}
        \Xhline{0.5pt}
        \ & \multicolumn{6}{c}{\textbf{Task}}\\
        \cline{2-7}
         \multirow{2}{*}{\textbf{Model}} & \multicolumn{3}{c|}{\textbf{Short instructions}} & \multicolumn{3}{c}{\textbf{Long instructions}}\\
          & ESR(a) & ESR(b) & SSL & ESR(a) & ESR(b) & SSL\\
        \Xhline{0.3pt}
         Language-Planner \cite{LMPlanner}& 40\% & 71.67\% & 33.92\% & 25\% & 64.67\% & 43.61\% \\
         Robot-Prompt \cite{RobotPrompt}& 55\% & 78.33\% & 71.80\% & 40\% & 80.25\% & 73.59\%\\
         MEIA w/o MEM & 50\% & 70.00\% & 64.12\% & 15\% & 56.33\% & 47.13\%\\
         MEIA w/o Large Model* & 70\% & 79.16\% & 77.17\% & 55\% & 84.58\% & 78.51\%\\
         \Xhline{0.3pt}
         MEIA & \textbf{95\%} & \textbf{97.50\%} & \textbf{97.00\%} & \textbf{85\%} & \textbf{94.17\%} & \textbf{92.82\%}\\
        \Xhline{0.5pt}
    \end{tabular}
    }
    \vspace{-10pt}
    \caption{The result for instruction planning evaluation. ESR(a) is the ESR for instruction and ESR(b) is the ESR for sub-tasks in  instruction. For MEIA w/o Large Model, We replaced GPT4 with GPT3.5.}
        \vspace{-5pt}
    \label{table:ins_t}
\end{table*}

\textbf{Visual question answering.}  For the visual question answering scenario, we select the scene that the robot needs to identify the air conditioner controller and select the correct control button according to the instructions.

\textbf{Spatial understanding and reasoning.}  We select visual navigation tasks to assess the spatial understanding and reasoning ability. The robot needs to select the appropriate location in the image based on the scene and customer requirements. The evaluation results are shown in Table \ref{table:sub_t}.

MiniGPT-4 has strong language capabilities. It can effectively understand the content of the prompt and provide outputs that are basically in line with the format. Nevertheless, its visual capability is limited, posing challenges in comprehending low-resolution and intricate images, thereby hindering effective task completion. CogVLM’s visual ability is relatively prominent. While it can effectively identify various targets on complex images and have an appropriate understanding of them, its grasp of prompts is insufficient, leading to difficulties in generating the required responses. CogAgent, built on the CogVLM design, shows no significant difference in accuracy compared to CogVLM in our evaluations.
However, CogAgent’s answers are quite stronger in terms of executability than CogVLM. The performance of GPT4-V is significantly better than these open-source models, and it maintains extremely high performance in various tasks. This will also be the basis for our subsequent experiments.

\begin{figure}[!t]
	\centering
    \includegraphics[width=0.5\textwidth]{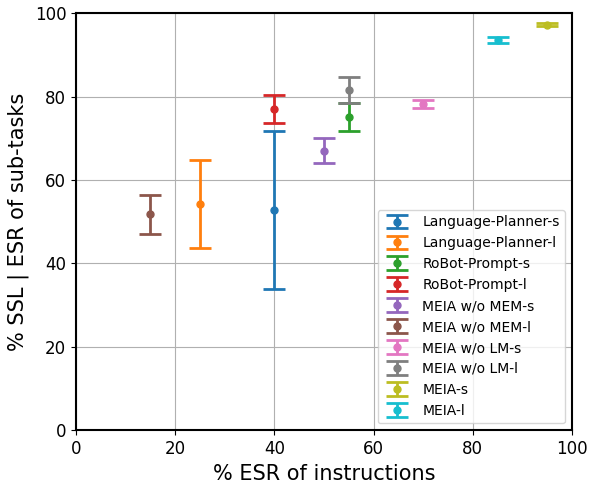}
	\caption{The result for instruction planning evaluation. -s for short instructions and -l for long instructions.
}
	\label{fig:plot}
\end{figure}

\subsubsection{Instruction Planning Evaluation}

To compare MEIA with existing methods, we design an instruction planning evaluation. We use \textbf{GPT3.5} to sample events and tasks in the cafe scenario and generate natural language instructions that include a certain number of sub-tasks, which are used as evaluation cases. To compare the effect of the number of sub-tasks in instructions on the model planning performance, we divide the generated instructions into two types: long instructions containing 3 to 5 sub-tasks and short instructions containing 2 to 3 sub-tasks. We generate 20 samples for each task and conduct experiments in the simulator to evaluate the models' \textbf{ESR} and \textbf{SSL} metrics. The instruction generation prompt and some of the evaluation examples and results can be found in the appendix.

\begin{figure*}[!t]
	\centering
    \includegraphics[width=1\textwidth]{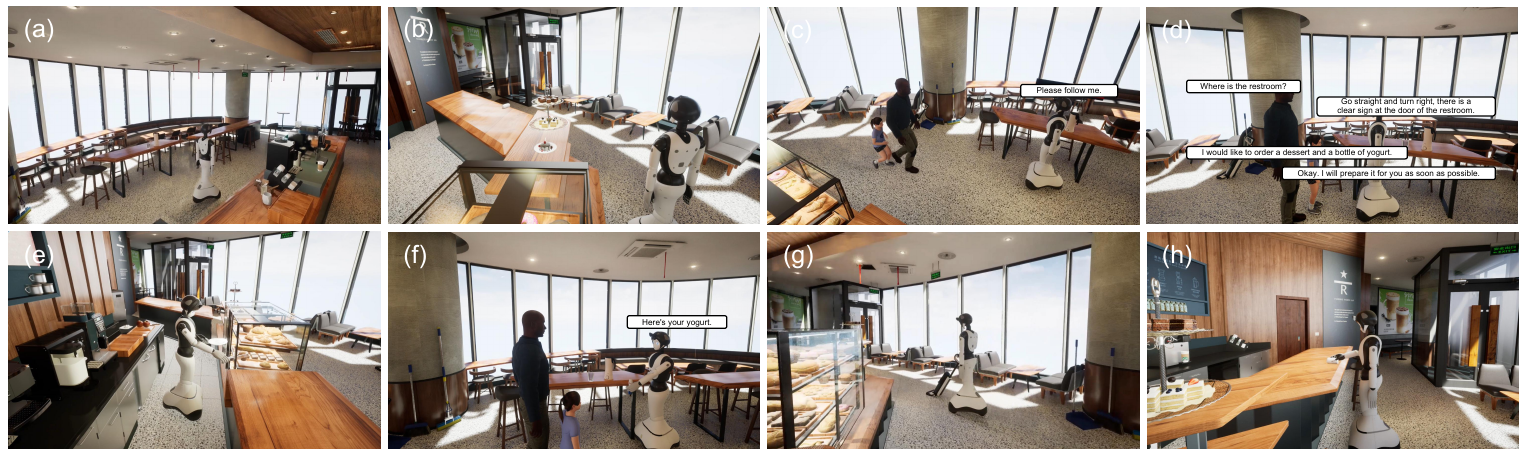}
\caption{\textbf{Full Pipeline Evaluation.} In (a) the agent explores the environment and constructs multimodal environmental memory. In (b) and (c), the agent waits for the customers and guides them to suitable seats. In (d) the agent conducts a conversation with customers and summarizes the conversation content to extract their needs. In (e) and (f) the agent plans and executes a sequence of actions to complete the order. In (g) and (h), the agent fulfills other tasks according to the instructions like cleaning the floors and wiping the table.
}
	\label{fig:experiment}
\end{figure*}

We evaluate the recent models Language-Planner \cite{LMPlanner} and Robot-Prompt \cite{RobotPrompt}. Language-Planner uses LLM for planning, adding the most relevant example from the example set along with the corresponding actions and the task to be completed as prompt, while Robot-Prompt uses the method of CoT to form high-performance action planning through multi-round dialogue and user feedback.
We make appropriate adjustments to the Language-Planner and Robot-Prompt based on the simulator environment, robot capabilities, and task requirements. The hyperparameter CUTOFF for Language-Planner is set to 0.6, and the number of examples is set to 10. We format the 3D environment of the cafe according to the requirements of Robot-Prompt. For more experimental results on hyperparameters and text format descriptions of the cafe environment, please refer to the appendix. The evaluation is conducted using samples generated by GPT3.5, as shown in Table \ref{table:ins_t}.

As shown in Table \ref{table:ins_t}, MEIA’s performance far exceeds that of Language-Planner and Robot-Prompt. We attribute it to our multimodal environment memory. Multimodal environment memory enables the model to have a more logical understanding of the environment, thereby ensuring the order and coherence of planning. This is reflected in the fact that the difference between MEIA’s ESR of instructions and that of other models is much greater than the difference in their ESRs of sub-tasks. The impact of instruction length on model performance is mainly reflected in the completion rate of instructions, and has little effect on the completion rate of sub-tasks and the accuracy of planning. Even with the increase of instructions, the model’s planning tends to be more stable, which leads to a slight increase in success rate. As shown in Fig.\ref{fig:plot}, the differences between ESR of instructions and SSL reflect the precision of model planning. Due to Language-Planner generating plans based on the similarity of examples and actions, its planning often results in confusion and redundancy, leading to particularly low SSL. Robot-Prompt lacks logical understanding of actions, causing it to occasionally lose key actions, resulting in a decrease in SSL.

\subsubsection{Full Pipeline Evaluation}

We evaluate the full plot pipeline. Since each evaluation contains more than ten sub-tasks, existing models or methods alone cannot complete the pipeline we design. Therefore, we conduct experiments with MEIA using large models such as GPT3.5, GPT4, and GPT4-V. The process of task execution can be seen in Fig.\ref{fig:experiment}. In multi-round experiments(for 20 rounds), MEIA’s ESR reaches \textbf{95\%}. This means that MEIA has extremely high accuracy and executability in planning.

We also study cases of failure. In one such case, the model makes a mistake in responding to “the table is dirty” by neglecting the initial step of "taking a towel" and directly going to the table, which reflects the large model’s insufficient understanding of the relation between embodied control ability and real-world scenarios.

\subsubsection{Ablation Study}
We conduct an ablation study on Multimodal Environment Memory(MEM) and Large Model of MEIA to validate their effectiveness. We conduct experiments using the same settings as in the instruction planning evaluation, and the results can be seen in Table \ref{table:ins_t} and Fig.\ref{fig:plot}.

We find that when we change the large model from GPT4 to GPT3.5, the performance of the model consistently decreases, reflecting the importance of the large model on instruction planning.
When we eliminate the multimodal environmental memory, the model's performance exhibits a more pronounced decline. Particularly concerning the ESR of instructions, it shows a greater regression compared to downgrading the large model. At the same time, the gap between the ESR of subtask and SSL has also widened significantly, reflecting a decrease in the precision of planning. We attribute it to the lack of multimodal environmental memory. The absence of multimodal environmental memory compels the large model to place more emphasis on the content of instructions. However, the precision of the instruction description cannot support the robot to successfully execute tasks, which affects the task execution success rate of the robot.

\subsection{Embodied Question Answering Evaluation}
In addition, we also evaluate the model on the embodied question answering task.

We randomly select 6 scenes from the dataset, with a total of 90 questions, and conduct experiments on the model in the form of single-round question answering and multi-round question answering.
In the single-round question answering task, the robot starts with an empty environment memory when receiving a question and explores from scratch each time.
In the multi-round question answering task, the environmental information explored to answer previous questions is available for generating answers to the current question.
Each set of question-answer pairs for each scene is considered as one testing round.

The experimental results are shown in the Table \ref{table:performance}.
Metrics used to evaluate the performance of the model include average accuracy, average exploration count, average unreachable planning count, and average path length. The \textbf{average accuracy(ACC)} measures the proportion of correct answers among multiple questions. The \textbf{average exploration count(EC)} counts the number of times the robot moves to the next exploration position planned by the model, excluding the initial exploration conducted by the robot at the starting position. The \textbf{average unreachable planning count(UPC)} tallies the number of positions in the environment that the model plans for the robot but are inaccessible. The \textbf{average path length(PL)} measures the total distance in centimeters that the robot navigates in the environment from receiving the question to generating the answer.

\begin{table}[!h]
\centering
\begin{tabular}{c|cccc}
\hline
   & ACC & EC & UPC & PL \\ \hline
single-round & 70\%  & 22     & 3.5       & 4769.9 \\ 
multi-round & 72\%  & 19.3   & 1         & 3800.7 \\ \hline
\end{tabular}
\caption{Comparison of results between single-round Q\&A and multiple-rounds Q\&A.}
\label{table:performance}
\end{table}

According to Table \ref{table:performance}, it can be observed that in the multi-round question answering task, the model achieves better results in terms of average exploration count, average unreachable planning count, and average path length while maintaining comparable or even improved average accuracy compared to the single-round question answering task. Particularly, the average unreachable planning count is reduced by more than 70\%. This result suggests the effectiveness of environment memory and historical collision information in the model's planning process. By leveraging the environment memory and historical collision information obtained through exploration, the model can better understand the overall environment, plan action, and expedite the exploration of the required environmental information for answering questions.

\subsubsection{Ablation Study} In order to further verify the effectiveness of the memory module, we present the results of the ablation experiment on the embodied question answering task. We test four models respectively: MEIA, MEIA without multimodal environment memory(MEM), MEIA without environmental language memory(ELM), and MEIA without environmental image memory(EIM). For each model, we calculate the average accuracy for different types of problems. The result is shown in Table \ref{table:ablation}.

\begin{table}[!h]
\centering
\scalebox{0.9}{\begin{tabular}{c|cccc}
\hline
      & location & comparing & existence & total   \\ \hline
MEIA w/o MEM & 38.9\%      & 50.0\%       & 77.8\%       & 56.7\% \\ 
MEIA w/o ELM & 41.7\%      & 50.0\%       & 77.8\%       & 57.8\% \\ 
MEIA w/o EIM & 50.0\%      & 55.6\%       & 80.6\%       & 63.3\% \\ \hline
\textbf{MEIA}    & \textbf{55.6\%}      & \textbf{66.7\%}      & \textbf{86.1\%}       & \textbf{70.0\%}  \\ \hline
\end{tabular}}
\caption{Comparison of average accuracy for different types of questions in the ablation experiment.}
\label{table:ablation}
\end{table}

Obviously, MEIA performs the best among the four models. The absence of either environmental language memory or environmental image memory has an impact on the model's performance. Furthermore, adding environmental image memory without environmental language memory does not significantly improve the model's performance.

It can also be observed from the result that the models perform the worst on location-type questions, followed by comparing-type questions. To some extent, it reflects the limited spatial reasoning ability of MEIA, and there is room for improvement in its overall performance.

\section{Conclusion}
In this work, we propose the MEIA for accomplishing embodied tasks and build an embodied question answering dataset with a thousand pieces of data. In the physical environment, the agent gathers egocentric environmental information, executes embodied control and completes embodied tasks. We innovatively introduce multi-modal memory to store global memory and update it in real time, thereby integrating embodied control with large models. Experiments indicate that MEIA demonstrates the promising ability to solve various embodied tasks.


\appendix
\bibliographystyle{named}
\bibliography{ijcai24}

\end{document}